\documentclass[conference]{IEEEtran}

\IEEEoverridecommandlockouts
\usepackage{cite}
\usepackage{amsmath,amssymb,amsfonts}
\usepackage{amsthm}
\usepackage{algorithmic}
\usepackage{graphicx}
\usepackage{textcomp}
\usepackage{xcolor}
\usepackage[]{footmisc}
\usepackage{bbm}

\usepackage{cuted}
\usepackage{stfloats}
\usepackage{mathtools,amssymb,lipsum, nccmath}
\newtheorem{theorem}{Theorem}

\usepackage{algorithm,algorithmic}
\usepackage{color,soul}
\usepackage{graphicx}
\usepackage{mathrsfs}
\usepackage{aligned-overset}
\usepackage{hyperref}
\usepackage{subcaption}
\newtheorem{Ass}{Assumption}

\def\BibTeX{{\rm B\kern-.05em{\sc i\kern-.025em b}\kern-.08em
    T\kern-.1667em\lower.7ex\hbox{E}\kern-.125emX}}

\begin{document}


\title{
Non-Convex Over-the-Air Heterogeneous Federated Learning: A Bias–Variance Trade-off
}
\author{Muhammad Faraz Ul Abrar, and Nicol\`{o} Michelusi~\IEEEmembership{Senior Member, IEEE}
\thanks{M. Faraz Ul Abrar and N. Michelusi are with the School of Electrical, Computer and Energy
Engineering, Arizona State University. email: \{mulabrar,
nicolo.michelusi\}@asu.edu.
This research has been funded in part by NSF under grant CNS-$2129015$.

An extended journal version of this paper is under review \cite{Biased_FL_Jorunal}.
}
}

\IEEEaftertitletext{\vspace{-0.25\baselineskip}}

\maketitle
\setulcolor{red}
\setul{red}{2pt}
\setstcolor{red}


\begin{abstract}
Over-the-air (OTA) federated learning (FL) has been well recognized as a scalable paradigm that exploits the waveform superposition of the wireless multiple-access channel to aggregate model updates simultaneously. Existing OTA-FL designs largely enforce zero-bias model updates by either assuming \emph{homogeneous} wireless conditions (equal path loss across devices) or forcing zero bias updates to guarantee convergence. Under \emph{heterogeneous} wireless scenarios, however, such unbiased designs are constrained by the worst-channel device and suffer from high variance in the updates. Moreover, prior analyses of biased OTA-FL largely focus on convex objectives, whereas most modern machine-learning models are highly non-convex. Motivated by these gaps, we study OTA-FL with stochastic gradient descent (SGD) for smooth non-convex objectives under wireless heterogeneity. We develop novel OTA-FL SGD updates that allow a structured, time-invariant model bias while facilitating reduced variance in the updates. We also derive a finite-time stationarity bound (expected time average squared gradient norm) that explicitly reveals a bias–variance trade-off. To optimize this trade-off, we pose a non-convex joint OTA power-control design and develop an efficient successive convex approximation (SCA) algorithm that requires only statistical CSI of the devices at the base station. Experiments on a non-convex image classification task validate the approach: the SCA-based design accelerates convergence via an optimized bias and improves generalization over prior OTA-FL baselines.
\end{abstract}


\vspace{-2mm}
\section{Introduction}
Federated Learning (FL) has emerged as a key paradigm for privacy-preserving distributed learning, allowing a massive number of devices to collaboratively train a model without sharing raw data. In a standard FL setup, $N$ clients with private datasets coordinate with a central parameter server (PS), e.g., an edge or cloud server, by transmitting only model parameters or gradients \cite{FL_Iot_Survey,FL_survey}. The objective is to solve
\begin{align}
\mathbf{w}^* = \arg\min_{\mathbf{w}\in\mathbb{R}^d} F(\mathbf{w}), \text{where } F(\mathbf{w})
\triangleq \frac{1}{N}\sum_{m\in[N]} f_m(\mathbf{w}), \tag{P}
\label{FL_prob}
\end{align}
$f_m(\mathbf{w})$ represents the local objective function of device $m$, and $F(\mathbf{w})$ is the global objective (loss) function. 
Problem \eqref{FL_prob} is typically solved by first-order optimization methods (e.g., mini-batch SGD): in each round, the PS broadcasts the current model $\mathbf{w}_t$, devices compute local gradients on their data and transmit them wirelessly to the PS. The PS then aggregates the received updates to obtain $\mathbf{w}_{t+1}$ and re-broadcasts it to begin the next round. This process repeats for multiple rounds until the global objective converges \cite{Fedavg}.

Nevertheless, realizing FL over wireless networks requires iterative exchange of high-dimensional parameters over noisy, bandwidth-constrained wireless channels, thereby posing a major performance bottleneck. A growing body of work, therefore, studies FL over wireless channels with robustness to fading and noise \cite{OTA_FL,FL_fading,One_bit_FL,Biased_OTA_FL_ICC}, and proposes device-scheduling strategies that account for channel gain fluctuations \cite{Sched_policies,ADFL,BB_FL}. Among these approaches, over-the-air (OTA) computation has emerged as a particularly scalable solution for wireless FL. OTA-FL works on the principle of waveform superposition on the multiple-access channel (MAC), whereby simultaneous transmissions yield ``single-shot" parameter aggregation at the PS, enabling fast FL updates \cite{OTA_FL,FL_fading,BB_FL}. To ensure unbiased estimation via OTA aggregation, the typical requirement is to align phases and equalize scales via device \emph{pre-scalers} and a PS \emph{post-scaler}. Over fading MACs, this coherent aggregation effectively requires channel inversion at the devices, and hence the feasible pre-scaling is dictated by the ``weakest" device. 

Thresholding and related heuristics have been proposed to mitigate this issue in OTA-FL literature \cite{FL_fading,BB_FL}. However, most prior OTA-FL analyses either assume \emph{wireless homogeneity} (equal average path loss across devices) to ensure zero average bias FL updates 
or explicitly enforce zero bias to facilitate convergence guarantees \cite{FL_fading,OTA_FL,One_bit_FL,OTA_FL_H_data}. In realistic heterogeneous deployments, where devices experience varying large-scale channel conditions, imposing zero bias can substantially increase model update variance (due to weak-channel devices). Conversely, relaxing the zero-bias design constraint, as in \cite{Opt_Power_Control_OTA_Comp,Opt_power_control_OTA_FL,OTA_FL_Optimization}, can introduce unstructured bias that is difficult to quantify, resulting in weaker convergence guarantees. In addition, much of the existing convergence theory for OTA-FL is developed under (strongly) convex objectives, see e.g.,\cite{OTA_FL_H_data,Opt_power_control_OTA_FL,Biased_OTA_FL_ICC}, which is inconsistent with the non-convex ML models deployed in modern practice.

To fill these gaps, we study OTA-FL under wireless heterogeneity with SGD updates for general smooth non-convex objectives. Building on and extending our prior work on biased OTA-FL in the strongly convex setting \cite{Biased_OTA_FL_ICC}, we design OTA-FL updates that admit a structured, time-invariant bias to reduce the variance of the updates, and we generalize the analysis to non-convex objectives. We establish a finite-time stationarity convergence bound (expected average squared gradient norm) that reveals a bias-variance trade-off. Building on this bound, we pose a joint OTA power-control problem and, unlike the heuristic designs proposed in \cite{Biased_OTA_FL_ICC}, develop an efficient successive convex approximation (SCA) algorithm that requires only statistical channel state information (CSI) at the PS, avoiding the heavy overhead of global instantaneous CSI. Finally, experiments on a non-convex learning task corroborate the theory, showing that the SCA-based design accelerates convergence by optimizing the allowed bias and achieves a target generalization performance faster than prior OTA-FL baselines.

\textit{Notation}: 
A zero-mean, circularly symmetric complex Gaussian random vector with mean $\mathbf m$ and covariance $\Sigma$ is denoted by $\mathcal{CN}(\mathbf m,\Sigma)$. $\Vert \cdot \Vert$ denotes the Euclidean $\ell$-2 norm. We let $[N]\equiv\{1,2,\dots,N\}$.
For a random vector $\mathbf v$, $\mathbb{E}[\mathbf v]$ denotes expectation and 
$\operatorname{var}(\mathbf v)\triangleq \mathbb{E}\!\left[\|\mathbf v-\mathbb{E}[\mathbf v]\|^{2}\right]$ denotes variance. 
\vspace{-2mm}
\section{System Model and over-the-air FL}
\begin{figure}[t!]
	\centering
	\includegraphics[width=0.35\textwidth]{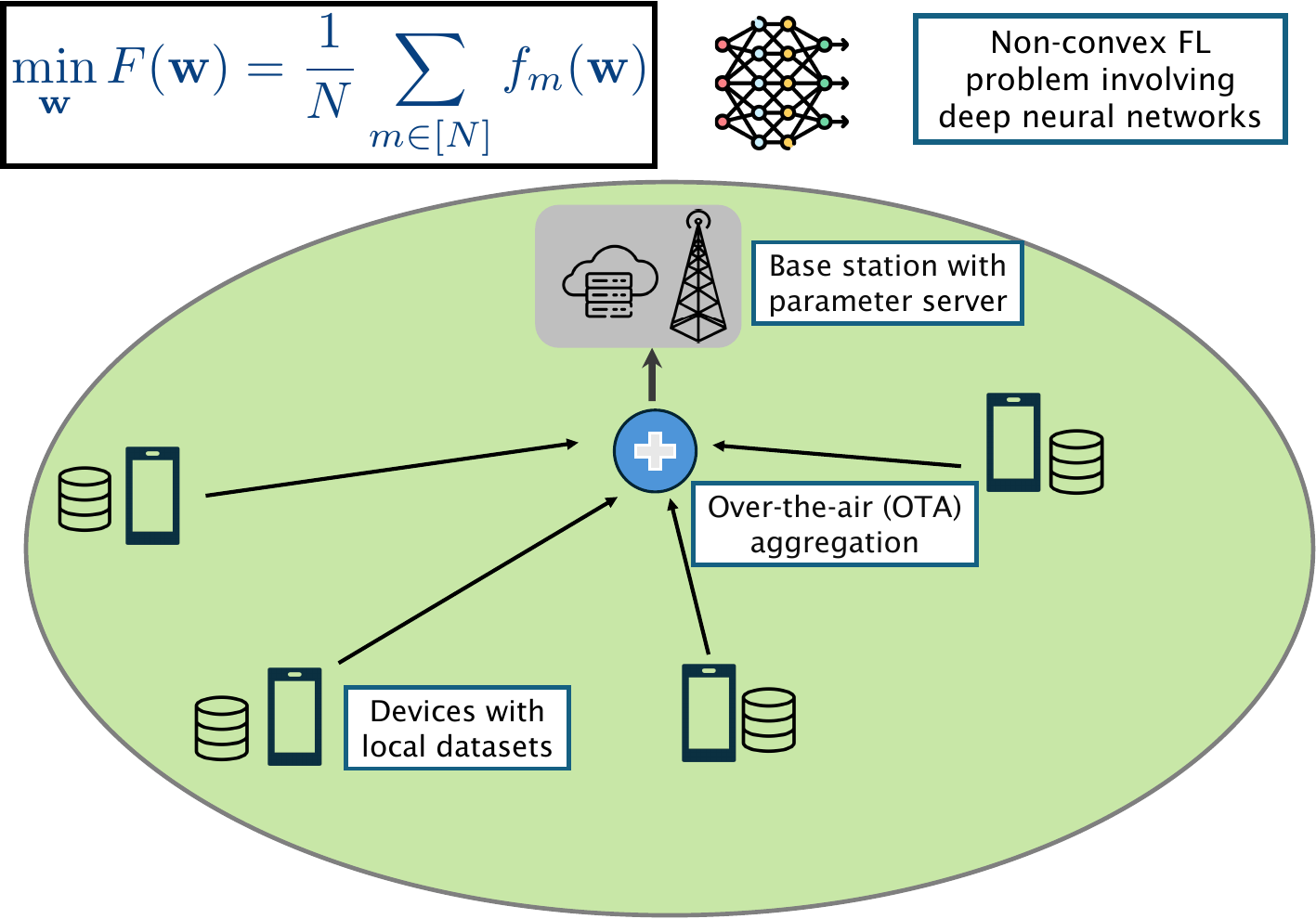}
	\caption{Non-convex OTA-FL in a wireless heterogeneous setup.\vspace{-5mm}}
\label{System_model}
\end{figure}

We consider a wireless network of $N$ distributed devices and a base station (also acting as the PS) aiming to collaboratively learn an FL model, as shown in Fig. \ref{System_model}. Each device $m$ owns a private local dataset $\mathcal{D}_m$ containing $D$ data points, with associated local objective function $f_m(\mathbf{w}) = \frac{1}{D} \,\sum_{\boldsymbol{\xi} \in \mathcal{D}_m} \phi(\mathbf{w},\boldsymbol{\xi})$, only computable at device $m$, where $\phi(\mathbf{w},\cdot)$ is the sample-wise loss function, $\boldsymbol{\xi}$ is a data point and $\mathbf{w} \in \mathbb{R}^d$ is the $d$-dimensional learning parameter. The goal is to solve \eqref{FL_prob} via distributed SGD over multiple FL rounds. Specifically, at the start of round $t$, the PS broadcasts the latest FL model $\mathbf{w}_t$ to each device in the network. Each device $m$ then uses a randomly drawn mini-batch  $\mathcal{B}_{m,t} \subseteq \mathcal{D}_m$ of fixed size $|\mathcal{B}_{m,t}| = B$ to estimate its local gradient $\boldsymbol{g}_{m,t} = \frac{1}{B} \sum_{\boldsymbol{\xi} \in \mathcal{B}_{m,t}} \nabla \phi(\mathbf{w}_t,\boldsymbol{\xi})$. Subsequently, each device uploads its estimated local gradient to the PS. Ideally, the PS aims to aggregate these gradients without error to compute the true global stochastic gradient:
\begin{align}
\overline{\boldsymbol{g}}_t = \frac{1}{N}\sum_{m \in [N]} \boldsymbol{g}_{m,t},
\label{Gradagg}
\end{align}
and subsequently updates the FL model with a fixed learning step size $\eta>0$ as
\begin{align}
	    \mathbf{w}_{t+1} = \mathbf{w}_{t} - {\eta} \overline{\boldsymbol{g}}_t, \quad t\geq0.  \label{GD}
\end{align}
This procedure is repeated until a target accuracy is reached or a fixed number of FL rounds is completed. Nonetheless, computing the global gradient in \eqref{Gradagg} requires noiseless aggregation of all the local gradients, i.e., they need to be perfectly received at the PS, and aggregated with uniform weights $\frac{1}{N}$. However, in practice, 
local gradients are obtained through a noisy wireless channel, introducing errors in the estimation of the global gradient $\overline{\boldsymbol{g}}_t$, as discussed next.

\subsection{Over-the-air transmission over a fading MAC}
We model the channel from device $m$ to the PS in round $t$ as flat Rayleigh fading, $h_{m,t}\sim\mathcal{CN}(0,\Lambda_m)$, independent across devices and i.i.d.\ over $t$. Here, the parameter $\Lambda_m$ represents the average 
channel gain, which is dependent on large-scale propagation conditions and is assumed to be constant throughout the FL runtime. We assume the PS has knowledge of this statistical CSI, i.e., $\{\Lambda_m\}$, but not the instantaneous CSI $\{h_{m,t}\}$. Critically, unlike existing works 
\cite{FL_fading,OTA_FL,One_bit_FL,OTA_FL_H_data} assuming identical average path loss across devices (${\Lambda_m{=}\Lambda_n , \forall m, n \in [N]}$), we consider a heterogeneous wireless environment where devices experience distinct path losses and study uplink transmission of local gradients via over-the-air (OTA) aggregation. OTA-FL exploits the superposition property of the wireless multiple-access channel to perform joint computation-and-communication, enabling single-shot aggregation at the PS. Following \cite{OTA_FL,FL_fading,BB_FL,One_bit_FL,NCOTA,OTA_FL_H_data}, we assume devices are perfectly synchronized during gradient upload. 
Denoting by $\mathbf{x}_{m,t}$ the signal transmitted by device $m$ in round $t$, the PS receives the signal 
\begin{align}
\mathbf{y}_t = \sum_{m \in [N]} h_{m,t} \cdot \mathbf{x}_{m,t} \;+    
\;\mathbf{z}_t,
\label{Signal_model}
\end{align}
where 
$\mathbf{z}_t \sim \mathcal{C N}(\mathbf{0},N_0 \mathbf{I})$
is the additive white Gaussian noise, i.i.d. over $t$. To approximate the ideal gradient aggregation in \eqref{Gradagg} using the signal model in \eqref{Signal_model}, each device pre-scales its local gradient using an OTA pre-scaler $\gamma_m$ and adopts a truncated channel inversion power control strategy, namely,
\begin{align}
\mathbf{x}_{m,t} =  
\frac{1}{ {h}_{m,t}}\chi_{m,t}\gamma_{m}\boldsymbol{g}_{m,t},
\label{OTA_signal_transmission}
\end{align}
where $\mathbf{\chi}_{m,t}$ is the OTA transmission indicator, defined as
\begin{align}
\mathbf{\chi}_{m,t} =  
\begin{cases}
1, \,\quad \text{ if } \vert h_{m,t}\vert \geq \frac{G_\text{max} \gamma_m}{\sqrt{d E_s}},\\
0,\quad \,\,\,\text{otherwise}.
\label{OTA_indicator}
\end{cases}
\end{align}
Here, $E_s$ is the maximum per-sample energy budget, and $G_\text{max}$ is an upper bound on $\Vert\boldsymbol{g}_{m,t}\Vert$ (see Assumption \ref{ass:bounded_loss_grad}). By construction, a device remains silent when $\vert h_{m,t}\vert < \frac{G_\text{max} \gamma_m}{\sqrt{d E_s}}$ to ensure the energy constraint $\Vert \mathbf{x}_{m,t}\Vert^2/d \leq E_s\,,\forall m,t$, where $d$ is the model dimension. The decision in~\eqref{OTA_indicator} is decentralized using the instantaneous uplink CSI $h_{m,t}$, which can be
acquired with negligible overhead via a downlink pilot, by leveraging channel reciprocity \cite{FL_d2d}. With this design, the PS estimates the global gradient \eqref{Gradagg} as
\begin{align}
\hat{\boldsymbol{g}}_t = \frac{\mathbf{y}_t} {\alpha}
=\frac{1}{\alpha} \sum_{m \in [N]}\chi_{m,t}\gamma_{m}\boldsymbol{g}_{m,t}
\;+    
\;\frac{\mathbf{z}_t}{\alpha},
\label{OTA_Signal_model2}
\end{align}
where $\alpha>0$ is a post-scaler, designed in the next section. Owing to concurrent uplink transmissions, the gradient upload time per round is independent of the number of active devices and equals $d/B$, where $B$ is the communication bandwidth.
\vspace{-1mm}
\subsection{Biased Over-the-Air-FL}
With the estimated gradient $\hat{\boldsymbol{g}}_t$ in \eqref{OTA_Signal_model2}, the PS updates the global model as
\begin{align}
 \mathbf{w}_{t+1} = \mathbf{w}_{t} - \eta \hat{\boldsymbol{g}}_t, \label{Prac_GD}
\end{align} in place of the ideal SGD update in \eqref{GD}.
Conditioned on $\mathbf{w}_t$ and the mini-batch selections, the received signal satisfies
$\mathbb{E}[\mathbf{y}_t]=\sum_{m\in[N]}\alpha_m\,\boldsymbol{g}_{m,t}$ with
$\alpha_m \triangleq \gamma_m\,\mathbb{E}[\chi_{m,t}]
= {\gamma_m \exp\{-\tfrac{\gamma_m^2 G_{\text{max}}^2}{d\,\Lambda_m\,E_s}\}}$, where the expectation is over channel fading and receiver noise. Choosing the post-scaler
$\alpha \triangleq \sum_{m\in[N]} \alpha_m$ yields a global gradient estimate such that the \emph{expected} gradient estimate is a convex combination of local stochastic gradients,
\begin{equation}
\tilde{\boldsymbol{g}}_t \;\triangleq\; \mathbb{E}[\hat{\boldsymbol{g}}_t \mid \mathbf{w}_t, \{\mathcal{B}_{m,t}\}_{m}]
\;=\; \sum_{m\in[N]} p_m\,\boldsymbol{g}_{m,t},
\label{exp_global_grad}
\end{equation}
with weights $p_m \triangleq \alpha_m/\alpha$, $0\le p_m\le 1$, and $\sum_{m}p_m=1$. We interpret $p_m$ as the
\emph{average participation level} of device $m$ induced by the OTA-FL design. Consequently, the update rule in \eqref{Prac_GD} acts as an SGD method that, on average, descends along $\tilde{\boldsymbol{g}}_t$ in place of the ideal aggregate $\overline{\boldsymbol{g}}_t$ from \eqref{Gradagg}. Therefore, these updates minimize a different objective function than \eqref{FL_prob}, on average, given by
 \begin{align}
    \tilde{F}(\mathbf{w}) = \sum_{m \in [N]} p_m f_m(\mathbf{w}).
    \tag{$\tilde{\text{P}}$}
    \label{incons_obj}
\end{align}
This can be readily verified by observing that $\mathbb{E}[\tilde{\boldsymbol{g}}_t ]= \nabla \tilde{F}(\mathbf{w}_t)$ with expectation taken over the mini-batch data selection.  

It is to be highlighted that prior OTA-FL works
\cite{Dig_vs_Analog,ADFL,FL_fading,OTA_FL,One_bit_FL,OTA_FL_H_data}
either assume homogeneous large-scale wireless conditions or enforce zero-bias aggregation, thereby ensuring uniform participation $ p_m = \frac{1}{N}$ for all $m \in [N] $, so that minimizing \eqref{incons_obj} becomes equivalent to \eqref{FL_prob}. While effective under homogeneous conditions, these designs become constrained by the weakest channel device in heterogeneous wireless networks, leading to high-variance model updates (see, e.g., \cite{BB_FL,Biased_OTA_FL_ICC,Opt_power_control_OTA_FL}). Moreover, under wireless heterogeneity, such schemes effectively minimize a \emph{different} objective, leading to objective inconsistency
\cite{Obj_incons} and a non-negligible model bias. As a result, their convergence guarantees do not directly apply to the scenario considered in this paper.

Our proposed OTA-FL aggregation design generalizes existing schemes by introducing a controllable, non-zero average bias, thereby subsuming zero-bias designs ($p_m=\tfrac{1}{N}$) as a special case. This formulation exposes an explicit
bias–variance trade-off that can be optimized to improve convergence. While some recent works \cite{Opt_power_control_OTA_FL,OTA_FL_Optimization} also consider biased OTA-FL, they typically allow a generic, unstructured bias that is difficult to analyze and control. In contrast, building on our prior work \cite{Biased_OTA_FL_ICC}, we adopt a structured, time-invariant bias that admits tractable finite-time guarantees. Note that, unlike our prior study \cite{Biased_OTA_FL_ICC}, which treated (strongly) convex objectives and offered heuristic designs, here we derive a convergence bound for non-convex objectives and optimize the bias-variance trade-off via an efficient SCA method, which is done in the next section.
\vspace{-0.8mm}
\section{Convergence Analysis and pre-scaler design}
In this section, we theoretically characterize the learning performance of the proposed OTA-FL scheme as governed by the choice of the device pre-scalers
$\{\gamma_m\}$. In the \emph{non-convex} setting considered here, we study convergence in terms of stationarity. To this end, our metric of interest is the expected (finite-time) average squared gradient norm of the global objective, given by $\frac{1}{T}\sum_{t=0}^{T-1}\mathbb{E}[\left\|\nabla F(\mathbf{w}_t)\right\|^2]$.
To study the convergence, we make the following standard assumptions:
\addtolength{\textheight}{0.04in}
\addtolength{\topmargin}{0.04in}
\begin{Ass}
\label{ass:smooth_lb}
For each device $m\in[N]$, the local objective $f_m(\cdot)$ is $L$-smooth, i.e.,
$$
\|\nabla f_m(\mathbf{a})-\nabla f_m(\mathbf{b})\|\le L\|\mathbf{a}-\mathbf{b}\|,\quad \forall\,\mathbf{a},\mathbf{b}\in\mathbb{R}^d,
$$
and is lower bounded, that is, there exists $f_m^{\inf}\in\mathbb{R}$ such that $f_m(\mathbf{w})\ge f_m^{\inf}\,,\forall \mathbf{w}\in\mathbb{R}^d$. 
Consequently, any convex combination $\sum_{m\in[N]} p_m f_m(\cdot)$ (including $F(\cdot)$ with $p_m=1/N$ and $\tilde F(\cdot)$) is $L$-smooth and lower bounded by $\sum_{m\in[N]} p_m f_m^{\inf}$.
\end{Ass}
\begin{Ass}  
\label{ass:bounded_loss_grad}
The sample-wise loss gradient for any given individual data sample $\xi$ is uniformly bounded, i.e., $\Vert \nabla \phi(\mathbf{w}, \xi) \Vert \leq G_\text{max},\forall \;\mathbf{w}\in \mathbb{R}^d$. It then follows from the triangle inequality that $\Vert\mathbf g_{m,t}\Vert\leq G_\text{max},\forall m,t$.
\end{Ass}  
\begin{Ass} 
\label{ass:bounded_stochastic_grad}
The mini-batch local gradient $\boldsymbol{g}_{m,t}$ is an unbiased estimate of the full-batch local gradient with bounded variance, i.e., $\mathbb{E}[\boldsymbol{g}_{m,t}|\mathbf w_t] = \nabla f_m(\mathbf{w}_t)$ and $
\mathrm{var}(\boldsymbol{g}_{m,t}|\mathbf w_t) \leq \sigma_m^2, \, \forall m \in [N], \, \mathbf{w}_t \in \mathbb{R}^d, t \geq 0$. \end{Ass}
\begin{Ass} 
\label{ass:bounded_data_hetr}
The variance of local gradients with respect to the global gradient is bounded, that is, there exists $\kappa >0 $ such that for all $\mathbf{w} \in \mathbb{R}^d\,,\,\frac{1}{N}\sum_{m \in [N]}
 \Vert \nabla f_m(\mathbf{w})-\nabla F(\mathbf{w})\Vert^2 \leq \kappa^2$. Under Assumption \ref{ass:bounded_loss_grad}, it further follows that $\kappa \leq 2 G_\text{max}$. \end{Ass}
Assumptions \ref{ass:smooth_lb} (smoothness) and \ref{ass:bounded_stochastic_grad} are standard in FL analyses (e.g., \cite{niid_fedavg,OTA_FL_H_data,OTA_FL_Optimization}), while Assumption \ref{ass:bounded_loss_grad} is also commonly adopted by many wireless FL works, see e.g., \cite{Transmission_Power_Control_OTA_FA,OTA_FL_H_data,OTA_FL_Optimization}.  The lower-boundedness in Assumption \ref{ass:smooth_lb} is mild and typical for non-convex objectives \cite{One_bit_FL,OTA_FL_Optimization}. Finally, Assumption \ref{ass:bounded_data_hetr}, often called bounded gradient dissimilarity or data divergence, captures data heterogeneity and is widely used in the literature (e.g., \cite{P_FL_bounded_niid}).

\subsection{Main Convergence Results}
Now, we are ready to present our main finite-time stationarity convergence result. Since the proposed OTA-FL updates in \eqref{Prac_GD} on average track the biased objective $\tilde{F}(\mathbf{w)}$ in \eqref{incons_obj}, we approach the analysis by splitting $\left\|\nabla F(\mathbf{w}_t)\right\|^2$ into: $\Vert \nabla \tilde F(\mathbf{w}_t)\Vert^2$ and  $ \|\nabla F(\mathbf{w}_t) - \nabla \tilde F(\mathbf{w}_t)\|^2$. 
With this, the convergence bound is then given as follows.
\begin{theorem}
\label{thm:main_convergence}
Under Assumptions 1-4 and  a fixed learning step size $0 < \eta \le 1/L$, after $T$ FL rounds it holds that
\begin{align}
    \frac{1}{T}\sum_{t=0}^{T-1}\mathbb{E}\big[\|\nabla F(\mathbf{w}_t)\|^2\big]
    \le & \frac{4\,\max_{m\in[N]} (f_m(\mathbf{w}_0) - f_m^{\inf})}{\eta\,T} \nonumber \\ +  2 \eta L \zeta 
     + & 2N\kappa^2 \sum_{m \in [N]}\Big(p_m-\tfrac1N\Big)^2,
    \label{eq:convergence_bound}
\end{align}
where $\zeta$ is the gradient estimation variance, bounded as:
\begin{align}
    \zeta \!\triangleq & G_{\max}^2\underbrace{\sum_{m\in[N]}  \left(\frac{p_m\gamma_m}{\alpha}-p_m^2\right)}_{\text{transmission variance}}
    + \underbrace{\sum_{m\in[N]} p_m^2 \sigma_m^2}_{\text{mini-batch variance}}
    & \!+ \!\underbrace{\frac{d N_0}{\alpha^2}}_{\text{receiver noise}}.
    \label{eq:variance_bound}
\end{align}
\end{theorem}
\noindent The proof sketch of Theorem \ref{thm:main_convergence} is provided in the Appendix.

Theorem \ref{thm:main_convergence} via \eqref{eq:convergence_bound} and \eqref{eq:variance_bound} characterizes the convergence behavior of the proposed OTA-FL scheme through three key components: (1) an optimization term which depends on the initial FL model $\mathbf{w}_0$ that decays as $O(1/(\eta T))$; (2) a
noisy global gradient estimation variance term; and (3) a model bias term induced by possibly non-uniform average participation. The variance term $\zeta$
splits further into: (a) transmission variance arising from truncated channel inversion, and intermittent gradient transmissions caused by channel fluctuations, which randomizes device participation at each round; (b) mini-batch variance from stochastic mini-batch gradients; and (c) additive receiver noise. Importantly, the choice of pre-scalers $\{\gamma_m\}$ affects all three key components in the variance term as well as governs the resultant average model bias, critically revealing the fundamental bias-variance trade-off at the heart of this work. In particular, a smaller $\gamma_m$ choice lowers transmission variance and the bias term but also reduces $\alpha$, amplifying the receiver-noise term. Conversely, increasing $\gamma_m$ can enlarge $\alpha$ and suppress receiver noise, but shifts $\{p_m\}$ away from uniform, introducing a non-zero model bias. This trade-off calls for a careful optimization of the design parameters, which we develop in the next section. 
\vspace{-1mm}
\subsection{OTA pre-scalers design}
Here, we study the OTA-FL power control strategy to accelerate convergence. To this end, we consider the following pre-scaler design problem:
\begin{align}
    \min\limits_{\{\gamma_m\},\gamma_m > 0\,, m\in [N]}  2 \eta L \zeta +2N\kappa^2 \sum_{m \in [N]}\Big(p_m-\tfrac1N\Big)^2,
    \label{OTA_design} \tag{P1}
\end{align}
corresponding to the convergence bound in Theorem \ref{thm:main_convergence} excluding the initialization term, which is independent of the device pre-scalers $\{\gamma_m\}$. One can verify that due to the complicated dependence of bias and variance terms on the choice of pre-scalers $\{\gamma_m\}$, \eqref{OTA_design} is a non-convex optimization. Leveraging variable dependencies, we equivalently rewrite the problem as a minimization over the coupled variables $\mathcal{X} \triangleq \{\{\gamma_m\}, \{p_m\}, \alpha\}$, subject to the following additional coupling constraints: (i) $\gamma_m \exp\{-\tfrac{\gamma_m^2 G_{\max}^2}{d\Lambda_m E_s}\}=\alpha p_m$ (i.e., $\alpha_m=\alpha p_m$);
(ii) $0<\gamma_m\le \gamma_{m,\max}$, with $\gamma_{m,\max}=\sqrt{\tfrac{d\Lambda_m E_s}{2G_{\max}^2}}$;
(iii) $0\le \alpha \le \min_{n}\frac{\alpha_{n,\max}}{p_n}$, with $\alpha_{m,\max}=\sqrt{\tfrac{d\Lambda_m E_s}{2e\,G_{\max}^2}}$;
(iv) $p_m\in[0,1]$ and $\sum_{m \in [N]} p_m=1$ (probability simplex). 
Since $\alpha_m(\gamma_m)$ is quasi-concave and attains its maximum $\alpha_{m,\max}$ at $\gamma_{m,\max}$, the constraint (iii) enforces feasibility of (i) by ensuring $\alpha p_m=\alpha_m\le \alpha_{m,\max}$ for all $m$. Moreover, for fixed $(\alpha,p_m)$ the transmission-variance term increases with $\gamma_m$, and (i) admits two roots $\gamma_{m,1}\le \gamma_{m,\max}\le \gamma_{m,2}$, therefore the constraint (ii) causes no loss of optimality. Notably, the resulting problem remains non-convex due to non-convex terms in both the objective and constraints. To address this, we adopt a successive convex approximation (SCA) approach, which iteratively solves convex surrogates of the original problem by linearizing the non-convex components around the current iterate. This procedure is guaranteed to converge to a stationary point of the original non-convex problem \cite{Inner_approx}.

To employ the SCA procedure, at iteration $k$, given anchors $\{\bar\gamma_m\}$, $\{\bar p_m\}$, $\bar\alpha$, we:
(i) introduce epigraph variables $z_m$ and convexify
the first component in the transmission variance \eqref{eq:variance_bound} as
$\frac{p_m\gamma_m}{\alpha}\le z_m$ via a log-linear surrogate (see \eqref{OTA_z_epi});
(ii) linearize the second component in the transmission variance
$-p_m^2$ around $\bar p_m$ in the objective (see \eqref{OTA_objective_cvx});
(iii) relax and convexify $\gamma_m e^{-\frac{\gamma_m^2G_{\max}^2}{d\Lambda_m E_s}}=\alpha p_m$ by taking logs and first-order expansions at $(\bar\alpha,\bar p_m)$ (see \eqref{OTA_alpha_m_conv});
(iv) convexify the bound on $\alpha$ using a first-order approximation of $1/\alpha$ at $\bar\alpha$ (see \eqref{OTA_alpha_max_conv}).
This yields the convex subproblem at iteration $k$:
\begin{subequations}
\begin{align}
\nonumber
 &\min_{\{\gamma_m\},\{p_m\},\{z_m\},\alpha}
 \ \eta L \Big( G_{\max}^2\sum_{m} z_m + \tfrac{dN_0}{\alpha^2} + \sum_m p_m^2\sigma_m^2
 \\&{-}G_{\max}^2\sum_m  \bar p_m(2p_m-\bar p_m)\Big)
 +N\kappa^2\sum_m \Big(p_m-\tfrac{1}{N}\Big)^2,
 \label{OTA_objective_cvx}\\
 &\text{s.t. }\forall m\in[N]:\nonumber\\
 &\quad \ln(\bar\gamma_m\bar p_m)+\tfrac{\gamma_m}{\bar\gamma_m}+\tfrac{p_m}{\bar p_m}-2 \;\le\; \ln z_m+\ln \alpha, \label{OTA_z_epi}\\
 &\quad \ln(\bar\alpha\,\bar p_m)+\tfrac{\alpha}{\bar\alpha}+\tfrac{p_m}{\bar p_m}-2 \;\le\; \ln\gamma_m - \tfrac{\gamma_m^2 G_{\max}^2}{d\Lambda_m E_s}, \label{OTA_alpha_m_conv}\\
 &\quad 0\le \gamma_m \le \gamma_{m,\max},\quad
       \frac{p_m}{\alpha_{m,\max}} \le \frac{2\bar\alpha-\alpha}{\bar\alpha^{2}},\ \alpha\ge 0, \label{OTA_alpha_max_conv}\\
 &\quad 0\le p_m\le 1,\ \sum_m p_m=1.
\end{align}
\label{OTA_convexified}
\end{subequations}
Each subproblem in \eqref{OTA_convexified} is convex and can be solved efficiently (e.g., via CVX \cite{cvx}). We initialize the anchors with a feasible point and iterate until the objective improvement falls below a tolerance threshold or a maximum number of iterations is reached, yielding a high-quality solution to the original non-convex problem.
\section{Numerical Results}
In this section, we evaluate the proposed SCA-optimized OTA-FL design. To this end, we study the popular handwritten digit classification problem in an FL setting on the widely used MNIST dataset \cite{MNIST}, which consists of C = 10 classes from ``0” to “9” with each image of size 28 x 28 pixels. We performed experiments on a fully connected neural network with one hidden layer, ReLU non-linear activations, and a hidden layer dimension of 1024, yielding a $d=814{,}090$-dimensional parameter vector $\mathbf{w}$.  The FL training uses an $\ell_2$-regularized cross-entropy loss at each device, with a regularization coefficient of 0.01. The setup results in a non-convex OTA-FL problem. We consider the FL problem with $N= 10$ devices uniformly deployed within a radius of $r_\text{max} = 1750 $ m
from the PS situated at the center. The communication bandwidth is $B$ = 1 MHz with carrier frequency $f_c = 2.4$ GHz, and the transmission power is set to $P_\text{tx} =$ 0 dBm. The noise power spectral density at the PS is $N_0 = -173$ dBm/Hz.
The large-scale channel condition $\Lambda_m$ follows the log-distance path loss model, with a path loss exponent 2.2 and 50 dB loss at the reference distance of $1$ m. 
To simulate a challenging FL scenario, we consider a heterogeneous (non-i.i.d.) data distribution. From a modified dataset of 10000 total samples (1000 per class), we adopt a data partition in which each device holds samples of exactly two digits (labels). Moreover, for any given label, its samples appear in the local datasets of no more than two devices. Each device computes full-batch gradients in every round, i.e., $|\mathcal{B}_{m,t}|=|\mathcal{D}_m|=1000$, which removes mini-batch variance in our simulations ($\sigma_m^2=0$ for all $m\in[N]$).
\begin{figure*}[t]
\centering
\subfloat[\centering  Test accuracy over FL rounds.]{{\includegraphics[width=0.4\textwidth]{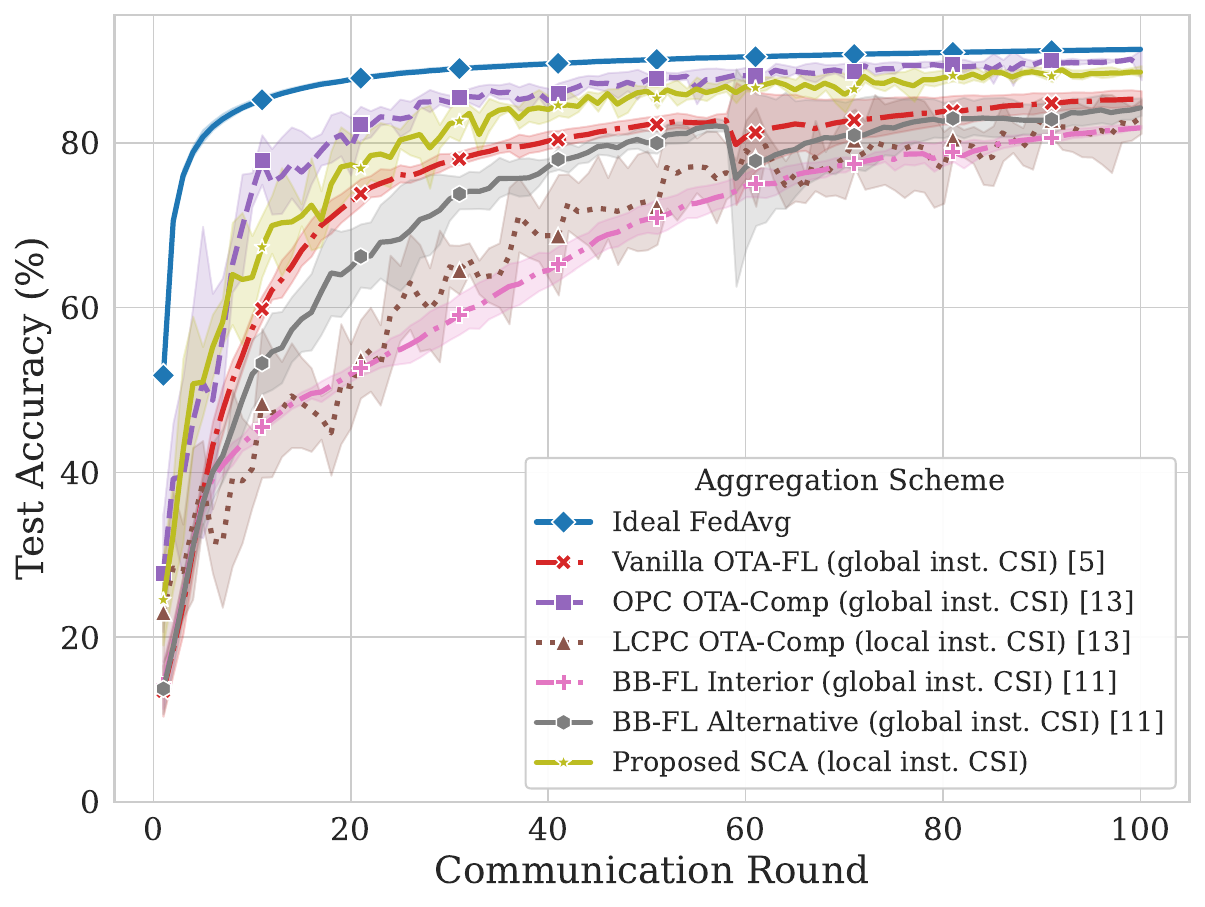} }}%
\qquad
\subfloat[\centering Global loss function $F(\mathbf{w})$ over FL rounds.]{{\includegraphics[width=0.4\textwidth]{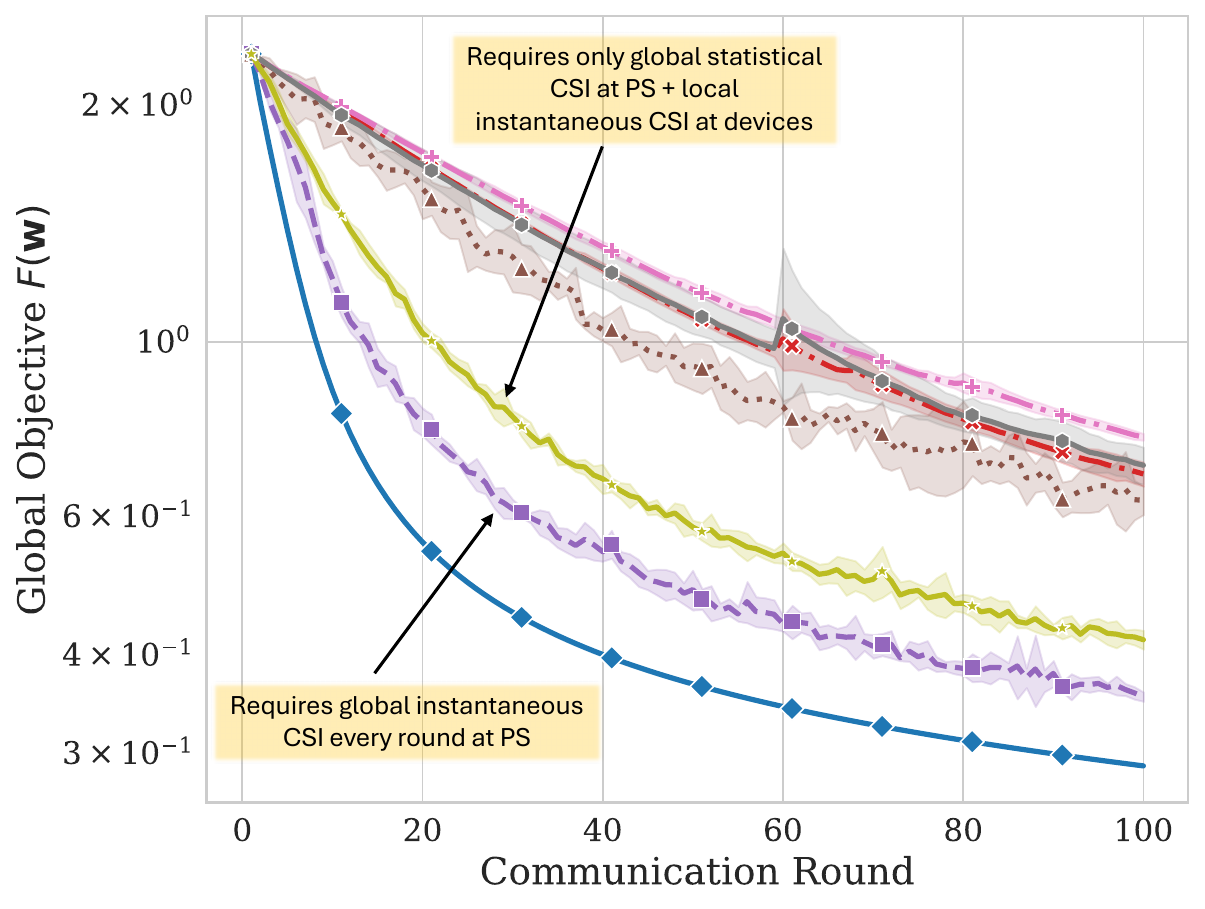} }}%
\caption{Comparison of various OTA-FL schemes, $N=10$ devices. The shared legend in (a) indicates each method’s CSI requirement at the PS for OTA power control design. }
\label{FL_results}
\end{figure*}
We evaluate the performance of the proposed SCA-optimized OTA-FL design against five OTA-FL baselines. From \cite{Opt_Power_Control_OTA_Comp}, we include Optimized Power Control (OPC) OTA-Comp, which minimizes the per-round mean squared error (MSE) using global instantaneous CSI, and its low-complexity variant, LCPC OTA-Comp, which uses truncated channel inversion with a common pre-scaler optimized with statistical CSI. We also compare against the classical Vanilla OTA-FL \cite{OTA_FL}, which enforces zero instantaneous bias via channel inversion, requiring global instantaneous CSI. Finally, we include two heterogeneity-aware schedulers from \cite{BB_FL}: BB-FL Interior, which schedules only devices within a fixed radius $R_\text{in}$, and BB-FL Alternative, which randomly alternates between full and interior scheduling. Acquiring global instantaneous CSI at the PS, as required by OPC OTA-Comp, Vanilla OTA-FL, and the BB-FL schemes, incurs substantial overhead. In contrast, our SCA design uses only statistical CSI at the PS together with local instantaneous CSI at the devices. We set $R_{\text{in}} = 0.6\,r_{\max}$ for BB-FL methods (as in \cite{BB_FL}), set $G_\text{max} = 10$, and select the best constant stepsize $\eta$ for all schemes via grid search. For reference, we also report Ideal FedAvg, which performs the noiseless update in \eqref{GD}.

Fig. \ref{FL_results} compares all methods over a fixed deployment, averaged across random fading and noise realizations. Fig. \ref{FL_results}a reports the test accuracy versus rounds, while Fig. \ref{FL_results}b shows the global objective. Ideal FedAvg demonstrates the best performance, since it aggregates gradients noiselessly. Among the practical wireless schemes, OPC OTA-Comp achieves the fastest convergence, closely followed by our proposed design, thanks to its optimized bias-variance trade-off. The strong performance of OPC OTA-Comp, however, comes at a substantial cost: it requires global instantaneous CSI at the PS in every round to optimize device pre-scalers. The proposed design, requiring only statistical CSI at the PS and local instantaneous CSI at devices, closely tracks OPC OTA-Comp in both metrics and attains competitive final accuracy, confirming that reducing model update variance while allowing a small average bias can yield performance gains. 

Compared with other baselines, the proposed design, despite lacking global CSI, consistently outperforms Vanilla OTA-FL, which enforces zero instantaneous bias at the cost of larger model update variance. While LCPC OTA-Comp also uses statistical CSI, it can introduce a less controllable bias, leading to noticeably noisier and slower convergence. Finally, the wireless heterogeneity-aware schedulers of \cite{BB_FL} behave as expected: BB-FL Alternative outperforms BB-FL Interior because restricting participation to a fixed interior subset reduces class coverage under our non-i.i.d. data split and weakens the model's generalization. Overall, by optimizing the bias-variance trade-off, the proposed design achieves faster loss decay and stronger generalization than the baselines without requiring global instantaneous CSI.


\section{Conclusion}
In this paper, we have studied non-convex OTA-FL under heterogeneous wireless conditions. Departing from conventional zero-bias model updates, we proposed an OTA aggregation scheme that introduces a structured, time-invariant model bias to mitigate the high update variance caused by channel disparities. We established convergence to a stationary point and derived a bound, which explicitly exposes a bias-variance trade-off driven by device power control design. Guided by this theoretical insight, we formulated a non-convex joint power-control design and developed an efficient SCA algorithm that requires only statistical CSI at the base station. Numerical results on a non-convex image classification task validated our theory, demonstrating that under wireless heterogeneity, allowing a controlled bias can noticeably accelerate model convergence and improve model generalization relative to state-of-the-art OTA-FL baselines.
\appendix
\noindent \textit{Proof sketch for \eqref{eq:convergence_bound}:} 

We start with $\nabla F(\mathbf{w}_t){=}\nabla \tilde F(\mathbf{w}_t){+}(\nabla F(\mathbf{w}_t){-}\nabla \tilde F(\mathbf{w}_t))$. Next, using $\Vert\mathbf a+\mathbf b\Vert^2
\leq
2\Vert\mathbf a\Vert^2
+2\Vert\mathbf b\Vert^2
$ for arbitrary vectors $\mathbf{a},\mathbf{b}$, it follows that
$\|\nabla F(\mathbf{w}_t)\|^2 \leq 2\|\nabla \tilde F(\mathbf{w}_t)\|^2
+2\|\nabla F(\mathbf{w}_t) - \nabla \tilde F(\mathbf{w}_t)\|^2.
$
Hence, taking expectations and averaging over $t$ yields $\frac{1}{T}\sum_{t=0}^{T-1}\mathbb{E}[\|\nabla F(\mathbf{w}_t)\|^2] $
\begin{align}
    &\leq \frac{2}{T}\sum_{t=0}^{T-1}\mathbb{E}[\|\nabla \tilde F(\mathbf{w}_t)\|^2]  + 
    \frac{2}{T}\sum_{t=0}^{T-1}\mathbb{E}[ \|\nabla F(\mathbf{w}_t) - \nabla \tilde F(\mathbf{w}_t)\|^2].
    \label{sub_optmality_decomposer}
\end{align}

First, we establish an upper bound on the first term of the right-hand side of \eqref{sub_optmality_decomposer}. Utilizing $L$-smoothness of $\tilde{F}(\cdot)$ (Assumption \ref{ass:smooth_lb}) at $\mathbf{w}_t$ and $\mathbf{w}_{t+1}$, we have $ \tilde F(\mathbf{w}_{t+1})$
\begin{align}
    \leq \tilde F(\mathbf{w}_{t}) \!+ \!\nabla \tilde{F}(\mathbf{w}_t)^T(\mathbf{w}_{t+1} \!- \!\mathbf{w}_t)\! +\! \frac{L}{2} \Vert\mathbf{w}_{t+1} - \mathbf{w}_t\Vert^2.
    \label{smoothness_F_tilde}
\end{align}
Recall that the model updates are given by $\mathbf{w}_{t+1} = \mathbf{w}_{t} - \eta \hat{\boldsymbol{g}}_t$, where $\hat{\boldsymbol{g}}_t$ is the estimate of the global gradient, with expected value $\mathbb E[\hat{\boldsymbol{g}}_t|\mathbf w_t]= \nabla \tilde F(\mathbf{w}_{t})$, and variance bounded by $\mathbb E[\Vert\hat{\boldsymbol{g}}_t- \nabla \tilde F(\mathbf{w}_{t})\Vert^2|\mathbf w_t]\leq \zeta$, where $\zeta$ is given in \eqref{eq:variance_bound}. The derivation of this variance bound (using Assumptions \ref{ass:bounded_loss_grad}–\ref{ass:bounded_stochastic_grad}) is omitted for brevity and can be found in our extended work \cite[Lemma 1]{Biased_FL_Jorunal}. It then follows that 
$\mathbb E[\mathbf{w}_{t+1} \!- \!\mathbf{w}_t|\mathbf w_t]=-\eta\nabla \tilde F(\mathbf{w}_{t})$
and
$\mathbb E[ \|\mathbf{w}_{t+1} \!- \!\mathbf{w}_t\|^2|\mathbf w_t]
\leq
\eta^2\Vert\nabla \tilde F(\mathbf{w}_{t})\Vert^2+\eta^2\zeta$. Applying expectation conditional on $\mathbf{w}_t$ to both sides on \eqref{smoothness_F_tilde}, we get
$$\mathbb{E}[\tilde F(\mathbf{w}_{t+1})|\mathbf w_t] \leq \tilde F(\mathbf{w}_{t}) - \frac{\eta}{2} \Vert \nabla \tilde{F}(\mathbf{w}_t)\Vert^2 + \frac{\eta^2 L}{2} \zeta,$$ where we utilized the fact that $\eta \leq 1/L$. Rearranging the above inequality, 
taking total expectations, summing from $t=0$ to $T-1$ and telescoping yields $\frac{2}{T}\sum_{t=0}^{T-1}\mathbb{E}\big[\|\nabla \tilde F(\mathbf w_t)\|^2\big]
\;\le\;
4\tfrac{\tilde F(\mathbf w_0)-\mathbb E[\tilde F(\mathbf w_T)]}{\eta T}
+2\eta L\,\zeta$.
By Assumption \ref{ass:smooth_lb}, $\tilde F(\mathbf w_T)\ge \sum_{m} p_mf_m^{\inf}$ and hence $\tilde F(\mathbf w_0)-\mathbb E[\tilde F(\mathbf w_T)]\le \sum_m p_m(f_m(\mathbf w_0)-f_m^{\inf})\le \max_{m}(f_m(\mathbf w_0)-f_m^{\inf})$ further results in $\frac{2}{T}\sum_{t=0}^{T-1}\mathbb{E}\big[\|\nabla \tilde F(\mathbf w_t)\|^2\big]
$
\begin{align}
&\leq \frac{4\,\max_{m\in[N]} \big(f_m(\mathbf w_0)-f_m^{\inf}\big)}{\eta T}
+2\eta L\,\zeta.
\label{init_term}
\end{align}
Now, we proceed to establish a bound on the second term on the right-hand side in \eqref{sub_optmality_decomposer}.  By definition, for arbitrary $\mathbf{w}$, $\| \nabla F(\mathbf{w}) -  \nabla\tilde{F}({\mathbf{w}})\|^2 = \big\| \sum_{m \in [N]} (p_m - \frac{1}{N}) \nabla f_m(\mathbf{w}) \big\|^2 = \big\| \sum_{m \in [N]} (p_m - \frac{1}{N})(\nabla f_m(\mathbf{w})-\nabla F(\mathbf{w})) \big\|^2 \leq ( \sum_{m \in [N]} (p_m - \frac{1}{N})^2) \cdot \sum_{m \in [N]} 
 \Vert \nabla f_m(\mathbf{w})- \nabla F(\mathbf{w})\Vert^2$, where we used the Cauchy-Schwarz inequality. Applying the definition of $\kappa$, we further obtain $\| \nabla F(\mathbf{w}_t) -  \nabla\tilde{F}({\mathbf{w}_t})\|^2 \leq N \kappa^2 \sum_{m \in [N]}  (p_m - \frac{1}{N})^2$. As a result, the second term in \eqref{sub_optmality_decomposer} can be bounded as
 \begin{align}
      \frac{2}{T}\sum_{t=0}^{T-1}\mathbb{E}[ \|\nabla F(\mathbf{w}_t) - \nabla \tilde F(\mathbf{w}_t)\|^2] \leq 2 N \kappa^2 \!\!\sum_{m \in [N]}  \Big(p_m - \frac{1}{N}\Big)^2.
      \label{Bias_term}
 \end{align}
Theorem \ref{thm:main_convergence} then follows by combining \eqref{init_term} and \eqref{Bias_term} into \eqref{sub_optmality_decomposer}.

\bibliographystyle{IEEEtran} 
\bibliography{Refs} 


\end{document}